\documentclass[conference]{IEEEtran}

\usepackage[utf8]{inputenc}
\usepackage[T1]{fontenc}
\usepackage{graphicx} 
\usepackage{amsmath}  
\usepackage{amssymb}  
\usepackage{booktabs} 
\usepackage{array}    
\usepackage{cite}     
\usepackage{url}
\usepackage{float} 
\begin{document}

\title{Short-Window Sliding Learning for Real-Time Violence Detection via LLM-based Auto-Labeling}
\IEEEoverridecommandlockouts

\author{
    \IEEEauthorblockN{
        Seoik Jung*, \quad
        Taekyung Song**, \quad
        Yangro Lee**, \quad
        Sungjun Lee\textdagger
    }
    
    \IEEEauthorblockA{
        \\[0.5ex] 
        PIA-SPACE Inc. \\[1ex] 
        \texttt{si.jung@pia.space}*, \texttt{tg.song@pia.space}**, \texttt{yr.lee@pia.space}**, \texttt{sj.lee@pia.space}\textdagger
    }
    
    \thanks{\textdagger Corresponding author.}
}

\maketitle

\begin{abstract}
This paper proposes a Short-Window Sliding Learning framework for real-time violence detection in CCTV footages. Unlike conventional long-video training approaches, the proposed method divides videos into 1-2 second clips and applies Large Language Model (LLM)-based auto-caption labeling to construct fine-grained datasets. Each short clip fully utilizes all frames to preserve temporal continuity, enabling precise recognition of rapid violent events. Experiments demonstrate that the proposed method achieves 95.25\% accuracy on RWF-2000 and significantly improves performance on long videos (UCF-Crime: 83.25\%), confirming its strong generalization and real-time applicability in intelligent surveillance systems.
\end{abstract}


\section{Introduction}
Recently, video-based violence and abnormal behavior detection has been gaining attention as an essential core technology in fields such as public safety, smart cities, and intelligent surveillance \cite{sultani2018}. Especially in real-time CCTV environments, it is crucial to quickly and accurately recognize rapidly occurring violent situations or abnormal behaviors within a short time \cite{sultani2018, cheng2021}.

However, existing research has mainly focused on training models using long-duration video data, and this approach has limitations that do not fit the temporal characteristics of actual surveillance environments \cite{zhang2025}. Previous studies rely on frame sampling from long video units, failing to sufficiently reflect temporal continuity and contextual information, and are also unsuitable for real-time CCTV environments in terms of computational efficiency \cite{zhang2025}.

Therefore, this paper proposes a short-window sliding learning technique. The proposed method involves dividing existing long videos into short clips of 1-2 seconds and auto-labeling each clip using a Large Language Model (LLM: e.g., Gemini). The generated labels are then reviewed by humans to construct a high-quality, fine-grained action dataset. This allows for more detailed learning of events that occur over a short period, such as violence. Furthermore, this study minimizes information loss due to sampling by utilizing all possible frames within each short clip for learning. This addresses the problem of relying on traditional frame sampling and allows for a richer reflection of temporal features within short intervals.

This study moves beyond the existing learning paradigm centered on long videos and presents the possibility of short-window-centric learning suitable for real-time surveillance environments. It is expected that this can be applied to various video-based application fields in the future, such as real-time abnormal behavior monitoring, violence detection, and emergency response systems.

\begin{figure}[!t]
    \centering
    \includegraphics[width=\columnwidth]{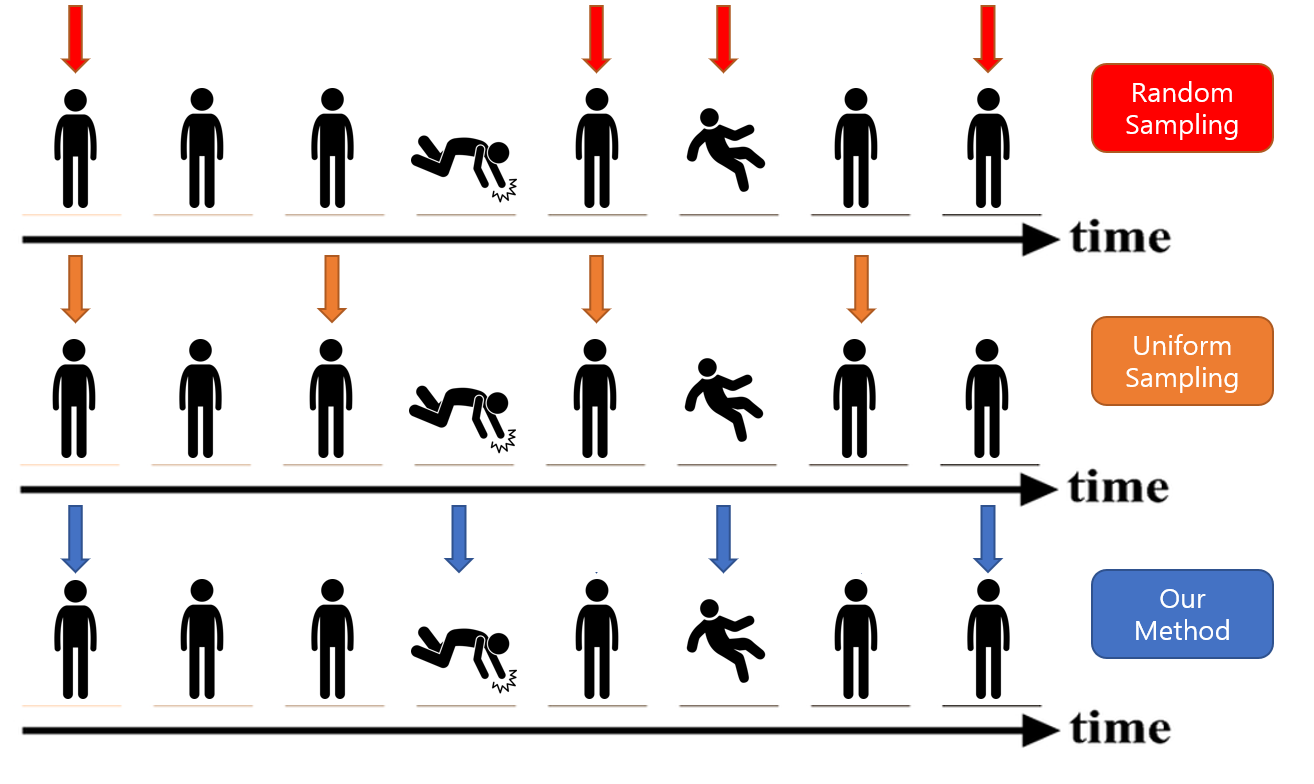}
    \caption{Comparison of frame sampling methods.}
    \label{fig:fig1}
\end{figure}

\section{Related Work}
This section summarizes the trends in video anomaly detection and violence recognition research and discusses the key differentiators of the proposed method from existing studies. Related work is examined in three categories: (1) video abnormal behavior detection, (2) violence video detection research, and (3) clip-level learning and auto-labeling approaches.

\subsection{Video Abnormal Behavior Detection}
Video abnormal behavior detection, the problem of distinguishing between normal and abnormal actions, has been studied using supervised, semi-supervised, and weakly supervised learning. Since Sultani et al.'s \cite{sultani2018} MIL (Multiple Instance Learning) framework, unsupervised learning using reconstruction-based and prediction-based approaches, as well as Transformer \cite{vaswani2017} models, has been actively pursued \cite{liu2025, abdalla2025}. Recently, research integrating temporal features through combination with vision-language models is expanding \cite{liu2024}. These studies mainly focus on detecting general abnormal behaviors such as theft, accidents, and congestion, and have limitations in precisely detecting behaviors that occur within a short time, like violence or attacks.

\subsection{Violence Video Detection}
Violence detection is a subfield of video anomaly detection, aiming to recognize aggressive acts such as physical conflict, striking, and chasing between people. Kaur et al. (2024) \cite{kaur2024} analyzed various violence recognition models and pointed out that most existing approaches are based on long-video-centric learning, limiting their ability to capture rapidly changing violent acts in a short time. Generally, violence detection models use long clips or sample a number of frames at regular intervals as input, making it difficult to sufficiently reflect temporal granularity \cite{jung2025}. Recently, multimodal approaches integrating visual and text information using vision-language models have been attempted, but most methods still rely on clip-level inputs and a frame-sampling-centric structure \cite{jung2025}.

\subsection{Clip-level Learning and Auto-Labeling}
Recently, attempts have been made to increase learning efficiency by dividing existing long video units into short clips or combining them with auto-labeling \cite{zhang2025}. Traditional sampling-based learning extracts frames at regular intervals from a long video, which can cause temporal information loss.

Consequently, most existing studies are based on long video inputs, sampling-centric processing, and clip-level labeling, lacking a structural approach for the detailed detection of violent situations occurring in brief moments. To solve these limitations, this study proposes a new learning framework that combines sub-second clip division, auto-caption labeling using LLMs, and the utilization of all frames within the clip. This method simultaneously improves temporal precision and real-time applicability, which were overlooked by existing methods.

\section{Methodology}
This study proposes a short-window sliding learning framework to precisely detect violent situations that occur in brief moments in real-time CCTV environments. The proposed method consists of two main stages: (1) data construction and (2) learning. Figure 2 illustrates the structure of the data construction process.

\subsection{Data Construction}
\textbf{Sliding window-based video segmentation.} Existing CCTV violence video data consists of clips 5-30 seconds long, with the actual violent act occurring in only a portion (about 1-2 seconds) of that time. Therefore, this study segments the original video using a sliding window approach with a set length of L seconds, window size W, and stride S.
$$ N = \frac{L-W}{S} + 1 $$
Here, N represents the total number of clips generated. Each clip maintains the temporal order of the original video, allowing the video's entire temporal information to be subdivided without loss.

\begin{figure}[!t]
    \centering
    \includegraphics[width=\columnwidth]{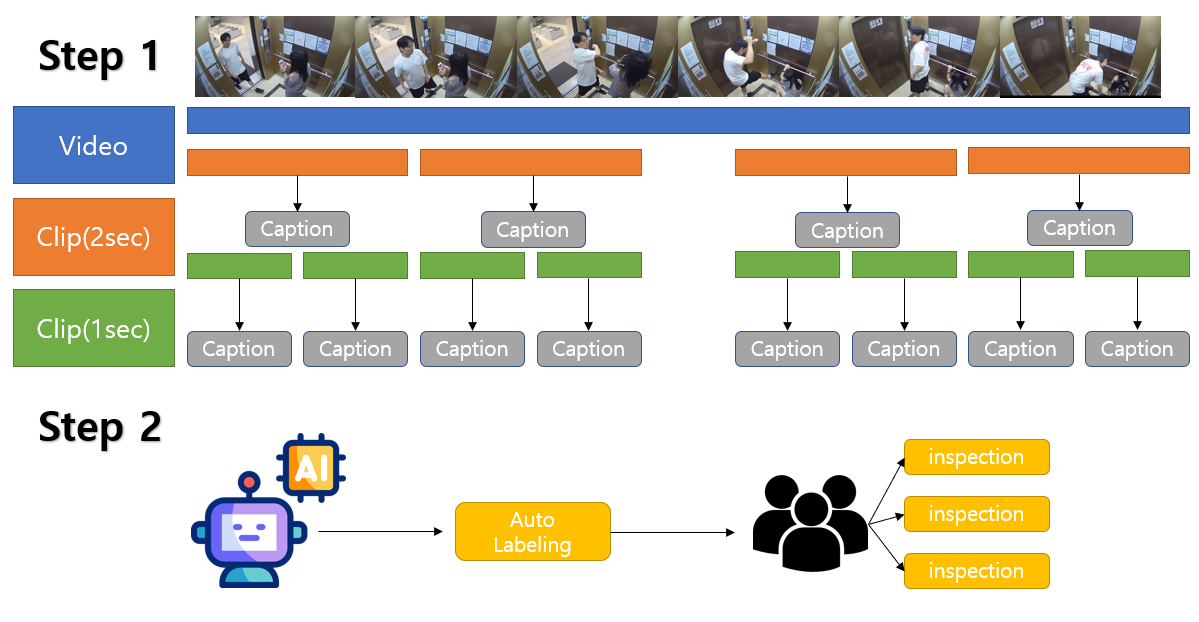}
    \caption{Proposed data construction process. Step 1 involves dividing the original video into 1-2 second clips and generating automatic captions. Step 2 builds a refined short video dataset through LLM-based auto-labeling and human review.}
    \label{fig:fig2}
\end{figure}

\textbf{Auto-labeling.} Captions were generated for each clip using Google's LLM service, Gemini. The model analyzes the main scenes of each clip and outputs sentence-form action descriptions such as "a scene of people pushing each other" or "an action of raising a fist". The generated captions are organized into a coarse-to-fine label structure. Subsequently, 3 reviewers manually checked the auto-generated captions to correct errors. The review work took about 40 hours per person, totaling 120 hours, and finally, a highly reliable short-video-based action dataset was constructed.
\begin{itemize}
    \item \textbf{Coarse Level:} Violence / Non-violence
    \item \textbf{Fine Level:} Punching, Kicking, Pushing, Chasing, etc.
\end{itemize}

\subsection{Learning and Experimental Setup}
This study used the InternVL3 \cite{zhu2025} architecture. This model is a VLM that processes video and text information integrally. It extracts features from each through a vision encoder and a language encoder, then aligns them in an intermediate representation space to learn semantic consistency.

By using all frames of a short interval (1-2 seconds) as input, temporal information loss that occurs in existing sampling-based learning was minimized. Caption information generated via LLM was provided as labels, enabling the model to learn not only visual patterns but also the meaning of the actions. In the inference phase, clips of 15-frame intervals were input in a sliding window fashion to predict violence in real-time. This structure allows for more accurate capturing of violent situations that occur in brief moments.

\begin{table*}[!t] 
    \centering
    \caption{Evaluation results on RWF-2000 after training on different domain data.}
    \label{tab:table1}
    
    \begin{tabular}{lcccccc}
        \toprule
        Experiment & AI hub / Short & RWF-2000 / Short & SCVD / Short & UCF-Crime / long & UCF-Crime / Short & Acc. \\
        \midrule
        
        Exp.0 & & & & & & 78.50\% \\
        Exp.1 & \checkmark & & & & & 82.50\% \\
        Exp.2 & & \checkmark & & & & 91.78\% \\
        Exp.3 & & & \checkmark & & & 88.75\% \\
        Exp.4 & & & & \checkmark & & 55.75\% \\
        Exp.5 & & & & & \checkmark & 83.25\% \\
        \textbf{Exp.6} & \checkmark & \checkmark & \checkmark & & & \textbf{95.25\%} \\
        \bottomrule
    \end{tabular}
\end{table*}

\section{Experiments}
In this section, experiments were conducted by setting up various datasets and learning scenarios to verify the effectiveness of the proposed short-window sliding learning technique. All experiments were evaluated based on the RWF-2000 dataset \cite{cheng2021}, which consists of 2000 short videos (5-10 seconds long) including violent and non-violent situations filmed in real CCTV environments. The evaluation goals of this study are (1) to verify the effectiveness of the proposed short-window learning technique, and (2) to analyze the generalization performance when video datasets from different domains are combined.

\subsection{Model and Training Environment}
The base model is InternVL3 \cite{zhu2025} with a multimodal Transformer structure, and all experiments were conducted with the same hyperparameters (input 12-15 frames, AdamW).

\subsection{Dataset Configuration}
A total of five datasets were used. For comparative experiments, each dataset was either configured in short clip units (proposed method) or kept as existing long videos.
\begin{itemize}
    \item \textbf{RWF-2000 \cite{cheng2021}:} Violence/non-violence video dataset collected in real CCTV environments (short videos).
    \item \textbf{AI Hub CCTV Dataset:} Short video clip dataset based on indoor/outdoor surveillance situations.
    \item \textbf{SCVD Dataset:} CCTV short video dataset including pedestrians, crowds, and abnormal situations.
    \item \textbf{UCF-Crime (long):} General abnormal behavior detection dataset composed of long videos several minutes long.
    \item \textbf{UCF-Crime (short):} The version segmented into 1-2 second units and auto-labeled using this study's method.
\end{itemize}

\begin{table}[H] 
    \centering
    \caption{Comparison with existing violence detection models.}
    \label{tab:table2}
    \begin{tabular}{lcc}
        \toprule
        Model / Method & Year & Accuracy \\
        \midrule
        MSTFDet \cite{qi2025} & 2025 & 95.20\% \\
        CUE-Net \cite{senadeera2024} & 2024 & 94.00\% \\
        Violence 4D \cite{magdy2023} & 2023 & 94.67\% \\
        Structured Keypoint Pooling \cite{hachiuma2023} & 2023 & 93.40\% \\
        Skeleton+Change Detection \cite{garcia2023} & 2023 & 90.25\% \\
        Semi-Supervised Hard Attention (SSHA) \cite{mohammadi2023} & 2022 & 90.40\% \\
        TwoStreamSepConv-LSTM \cite{islam2021} & 2021 & 89.75\% \\
        Flow Gated Network \cite{cheng2021} & 2021 & 87.25\% \\
        \midrule
        \textbf{Our Method} & \textbf{2025} & \textbf{95.25\%} \\
        \bottomrule
    \end{tabular}
\end{table}
\subsection{Quantitative Comparison}
In Table 2, the proposed method achieved 95.25\% accuracy on RWF-2000, slightly surpassing MSTFDet (95.2\%) \cite{qi2025}. Achieving SOTA-level performance with only short clip learning, without skeleton or multi-stream structures, confirmed that the proposed strategy is effective for capturing temporal features.

As shown in Table 1, the model trained on long videos (UCF-Crime/Long, Exp.4) showed a low accuracy of 55.75\%, whereas the Short version, segmented into 1-2 second units (Exp.5), improved to 83.25\%. The model that combined short clip data such as AI Hub, SCVD, and RWF-2000 (Exp.6) recorded the highest performance at 95.25\%, confirming the generalization effect of short-window learning.

\section{Conclusion}
In this study, we proposed a short-window sliding learning technique for precisely detecting violent situations that occur over a short period in real-time surveillance environments. By dividing videos into 1-2 second clips and performing auto-labeling and caption-based data augmentation using LLM, we simultaneously improved temporal precision and data efficiency compared to existing methods.

Experimental results showed that the proposed method achieved 95.25\% accuracy on the RWF-2000 dataset, outperforming previous state-of-the-art research. Furthermore, performance was significantly improved when long video data was converted to short clip units for training, and generalization performance was maximized when short video data from various domains was combined. These results suggest that the method proposed in this study can be effectively applied to the problem of real-time violence detection in actual CCTV environments.

Future research plans to expand in the direction of integrating multiple modalities (voice, subtitles, action features) to develop an intelligent real-time surveillance model that can cover more complex abnormal situations (fear, theft, suicide attempts, etc.).

\section*{Acknowledgment}
This paper was conducted as part of the "Advancement and Overseas Expansion of VLM-based Automatic Anomaly Detection Real-time Video Analysis AI Solution" (Project No: PJT-25-031547), a research task of the "2025 Regional Digital Basic Fitness Support (Leading Enterprise Commercialization Support Project)," supervised by the National IT Industry Promotion Agency (NIPA) with support from the Ministry of Science and ICT.



\begin{thebibliography}{99}

\bibitem{sultani2018}
W. Sultani, C. Chen, and M. Shah, "Real-world anomaly detection in surveillance videos," in \textit{Proceedings of the IEEE conference on computer vision and pattern recognition}, 2018, pp. 6479-6488.

\bibitem{liu2025}
J. Liu, et al., "Networking systems for video anomaly detection: A tutorial and survey," \textit{ACM Computing Surveys}, 2025, 57.10: 1-37.

\bibitem{abdalla2025}
M. Abdalla, et al., "Video anomaly detection in 10 years: A survey and outlook," \textit{Neural Computing and Applications}, 2025, 1-44.

\bibitem{liu2024}
Y. Liu, et al., "Generalized video anomaly event detection: Systematic taxonomy and comparison of deep models," \textit{ACM Computing Surveys}, 2024, 56.7: 1-38.

\bibitem{kaur2024}
G. Kaur and S. Singh, "Revisiting vision-based violence detection in videos: A critical analysis," \textit{Neurocomputing}, 2024, 597: 128113.

\bibitem{jung2025}
S. Jung, et al., "DUAL-VAD: Dual Benchmarks and Anomaly-Focused Sampling for Video Anomaly Detection," \textit{arXiv preprint arXiv:2509.11605}, 2025.

\bibitem{zhang2025}
H. Zhang, et al., "Holmes-vau: Towards long-term video anomaly understanding at any granularity," in \textit{Proceedings of the Computer Vision and Pattern Recognition Conference}, 2025, pp. 13843-13853.

\bibitem{qi2025}
B. Qi, B. Wu, and B. Sun, "Automated violence monitoring system for real-time fistfight detection using deep learning-based temporal action localization," \textit{Scientific Reports}, 2025, 15.1: 29497.

\bibitem{senadeera2024}
D. C. Senadeera, et al., "Cue-net: violence detection video analytics with spatial cropping enhanced uniformerv2 and modified efficient additive attention," in \textit{Proceedings of the IEEE/CVF Conference on Computer Vision and Pattern Recognition}, 2024, pp. 4888-4897.

\bibitem{magdy2023}
M. Magdy, M. W. Fakhr, and F. A. Maghraby, "Violence 4D: Violence detection in surveillance using 4D convolutional neural networks," \textit{IET Computer Vision}, 2023, 17.3: 282-294.

\bibitem{hachiuma2023}
R. Hachiuma, F. Sato, and T. Sekii, "Unified keypoint-based action recognition framework via structured keypoint pooling," in \textit{Proceedings of the IEEE/CVF conference on computer vision and pattern recognition}, 2023, pp. 22962-22971.

\bibitem{garcia2023}
G. Garcia-Cobo and J. C. Sanmiguel, "Human skeletons and change detection for efficient violence detection in surveillance videos," \textit{Computer Vision and Image Understanding}, 2023, 233: 103739.

\bibitem{mohammadi2023}
H. Mohammadi and E. Nazerfard, "Video violence recognition and localization using a semi-supervised hard attention model," \textit{Expert Systems with Applications}, 2023, 212: 118791.

\bibitem{islam2021}
Z. Islam, et al., "Efficient two-stream network for violence detection using separable convolutional lstm," in \textit{2021 International Joint Conference on Neural Networks (IJCNN)}, IEEE, 2021, pp. 1-8.

\bibitem{cheng2021}
M. Cheng, K. Cai, and M. Li, "RWF-2000: An open large scale video database for violence detection," in \textit{2020 25th International conference on pattern recognition (ICPR)}, IEEE, 2021, pp. 4183-4190.

\bibitem{vaswani2017}
A. Vaswani, et al., "Attention is all you need," \textit{Advances in neural information processing systems}, 2017, 30.

\bibitem{zhu2025}
J. Zhu, et al., "Internvl3: Exploring advanced training and test-time recipes for open-source multimodal models," \textit{arXiv preprint arXiv:2504.10479}, 2025.

\end{thebibliography}
\end{document}